\pdfoutput=1
\documentclass[11pt]{article}

\usepackage{EMNLP2023}

\usepackage{times}
\usepackage{latexsym}

\usepackage[T1]{fontenc}
\usepackage[utf8]{inputenc}

\usepackage{microtype}

\usepackage{inconsolata}
\usepackage{graphicx}
\usepackage{multirow}
\usepackage{booktabs}
\usepackage{amsmath}
\usepackage{amssymb}
\usepackage{subfigure}
\title{Unify word-level and span-level tasks: NJUNLP's Participation for the WMT2023 Quality Estimation Shared Task}

\author{Xiang Geng\textsuperscript{\rm 1}, Zhejian Lai\textsuperscript{\rm 1}, Yu Zhang\textsuperscript{\rm 1}, Shimin Tao\textsuperscript{\rm 2}, Hao Yang\textsuperscript{\rm 2}, Jiajun Chen\textsuperscript{\rm 1}, Shujian Huang\textsuperscript{\rm 1}\thanks{* Corresponding Author.}\\
\textsuperscript{ \rm 1} National Key Laboratory for Novel Software Technology, Nanjing University, Nanjing, China \\
\textsuperscript{ \rm 2 } Huawei Translation Services Center, Beijing, China\\
\texttt \{gx, laizj,  zhangy\}@smail.nju.edu.cn, \{taoshimin, yanghao30\}@huawei.com\\
\texttt \{chenjj, huangsj\}@nju.edu.cn
}
\begin{document}
\maketitle
\begin{abstract}
We introduce the submissions of the NJUNLP team to the WMT 2023 Quality Estimation (QE) shared task. 
Our team submitted predictions for the English-German language pair on all two sub-tasks: (i) sentence- and word-level quality prediction; and (ii) fine-grained error span detection. 
This year, we further explore pseudo data methods for QE based on NJUQE framework\footnote{\url{https://github.com/NJUNLP/njuqe}}.
We generate pseudo MQM data using parallel data from the WMT translation task.
We pre-train the XLMR large model on pseudo QE data, then fine-tune it on real QE data.
At both stages, we jointly learn sentence-level scores and word-level tags. 
Empirically, we conduct experiments to find the key hyper-parameters that improve the performance.
Technically, we propose a simple method that covert the word-level outputs to fine-grained error span results.
Overall, our models achieved the best results in English-German for both word-level and fine-grained error span detection sub-tasks by a considerable margin.
\end{abstract}

\section{Introduction}
Quality Estimation (QE) of Machine Translation (MT) is a task to estimate the quality of translations at run-time without access to reference translations~\cite{mtqe}. 
There are two sub-tasks in WMT 2023 QE shared task\footnote{\url{https://wmt-qe-task.github.io}}: (i) sentence- and word-level quality prediction; and (ii) fine-grained error span detection. We participated in all two sub-tasks for the English-German (EN-DE) language pair. The annotation of EN-DE is multi-dimensional quality metrics (MQM) \footnote{\url{https://themqm.org}}, aligned with the WMT 2023 Metrics shared task. 
The MQM annotation provides error spans with fine-grained categories and severities by human translators.

Inspired by DirectQE~\cite{directqe} and CLQE~\cite{clqe}, we further explore pseudo data methods for QE based on the NJUQE framework. 
We generate pseudo MQM data using parallel data from the WMT translation task. Specifically, we replace the reference tokens with these tokens sampled from translation models. 
To simulate translation errors with different severities, we sample tokens with lower generation probabilities for worse errors~\cite{geng2022njunlp}. 
We pre-train the XLMR~\cite{xlmr} large model on pseudo MQM data, then fine-tune it on real QE data. 
At both stages, we jointly learn sentence-level scores (MSE loss and margin ranking loss) and word-level tags (cross-entropy loss). 

For task (i), the QE model outputs the sentence scores and the ``OK'' probability of each token. For task (ii), we set different thresholds for the ``OK'' probability to predict fine-grained severities. We regard consecutive ``BAD'' tokens as a whole span and take the worse severity of each token as the result. We train different models with different parallel data and ensemble their results as the final submission.

Overall, we summarize our contribution as follows:
\begin{itemize}
    \item Empirically, we conduct experiments to find the key hyper-parameters that improve the performance.
    \item Technically, we propose a simple method that converts the word-level outputs to fine-grained error span results.
\end{itemize}

Our system obtains the best results in English-German for both word-level and fine-grained error span detection sub-tasks with an MCC of 29.7 (+4.1 than the second best system) and F1 score of 28.4 (+1.1) respectively. We rank 2nd place on sentence-level sub-tasks with a Spearman score of 47.9 (-0.4 than the best system).
\begin{table*}[t]
\centering
\resizebox{0.99\textwidth}{!}{
\begin{tabular}{l|l|l}
\hline
\textbf{Source} & \multicolumn{2}{l}{Government Retires 15 More Senior Tax Officials On Graft Charges } \\
\textbf{Translation} & \multicolumn{2}{l}{Regierung \textcolor{red}{zieht} 15 weitere leitende Steuerbeamte wegen \textcolor{red}{Graft-Vorwürfen} zurück} \\
\textbf{Translation Back} & \multicolumn{2}{l}{Government withdraws 15 more senior tax officials over graft allegations} \\
\hline
\hline
\textbf{Tags} & \multicolumn{2}{l}{OK \textcolor{red}{BAD} OK OK OK OK OK \textcolor{red}{BAD} OK}\\
\textbf{MQM Score} & \multicolumn{2}{l}{0.3333} \\
\hline
\hline
\textbf{Annotation ID} & \textbf{Character-level Indices of Error Span}  & \textbf{Severity} \\
\textbf{Span 1} & 10:15 & Major \\
\textbf{Span 2} & 55:70 & Minor \\
\hline

\end{tabular}
}
\caption{An example from the WMT2023 English-German MQM dataset. We mark the error span with red color. The translation back is generated by Google Translate.}
\label{tab:example}
\end{table*}
\section{Background}

\begin{figure*}[h]
\centering
\includegraphics[width=0.6\textwidth]{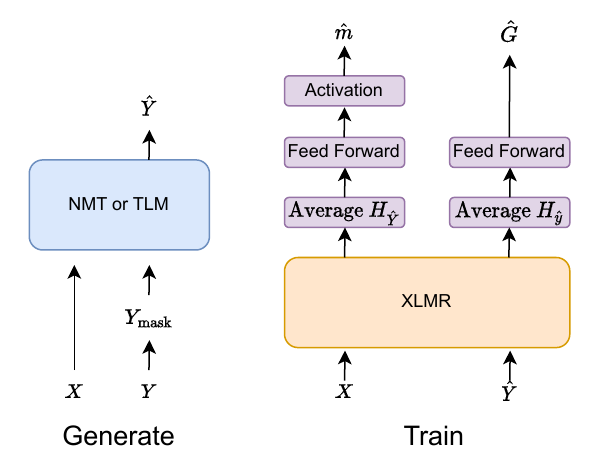}
\caption{
Illustration of the whole procedure.  
}
\label{fig:framework}
\end{figure*}

Given a source language sentence ${X}$ and a target language translation ${\hat{Y}} = \{y_1, y_2, \dots, y_n\}$ with $n$ tokens, the MQM annotation provides error spans with fine-grained categories and severities (minor, major, and critical) by human translators.
The MQM score sums penalties for each error severity and then normalizes the result by translation length:
\begin{align}\label{eq:mqm}
    \text{MQM} = 1 - \frac{n_{\text{minor}}+5n_{\text{major}}+10n_{\text{critical}}}{n},
\end{align}
where $n_{\text{severity}}$ denotes the number of each error severity and $n$ denotes the translation length.

As shown in table \ref{tab:example},  participating systems are required to predict tags ${G} = \{g_1, g_2, \dots, g_n\}$ of each word and MQM score $m$ for sub-task (i), where the binary label $g_j\in \{\text{OK}, \text{BAD}\}$ is the quality label for the word translation $y_j$. 
For sub-task (ii), we need to predict both the character-level start and end indices of every error span as well as the corresponding error severity. 
The primary metrics of sentence-level, word-level, and span detection sub-tasks are Spearman’s rank correlation coefficient, Matthews correlation
coefficient (MCC)\footnote{\url{https://github.com/sheffieldnlp/qe-eval-scripts/tree/master}}, and F1-score respectively\footnote{\url{https://github.com/WMT-QE-Task/wmt-qe-2023-data/blob/main/task_2/evaluation}}.

\section{Methodology}
Generally, we unite the sub-tasks (i) and (ii) as follows:
\begin{itemize}
    \item We generate pseudo MQM data for sub-task (i) using parallel data and translation models as shown in the left of figure \ref{fig:framework}.
    \item We pre-train the QE model with pseudo data and fine-tune it with real QE data for sub-task (i) as shown in the right of figure \ref{fig:framework}.
    \item We ensemble the results of models trained with different parallel data for sub-task (i).
    \item We convert word-level probabilities for sub-task (i) to error span and fine-grained severities for sub-task (ii).
\end{itemize}

\subsection{Pseudo MQM Data}
We adopt the pseudo MQM data method described in~\cite{geng2022njunlp}.
\subsubsection{Corrupting}
\begin{figure*}[t]
\centering
\includegraphics[width=0.8\textwidth]{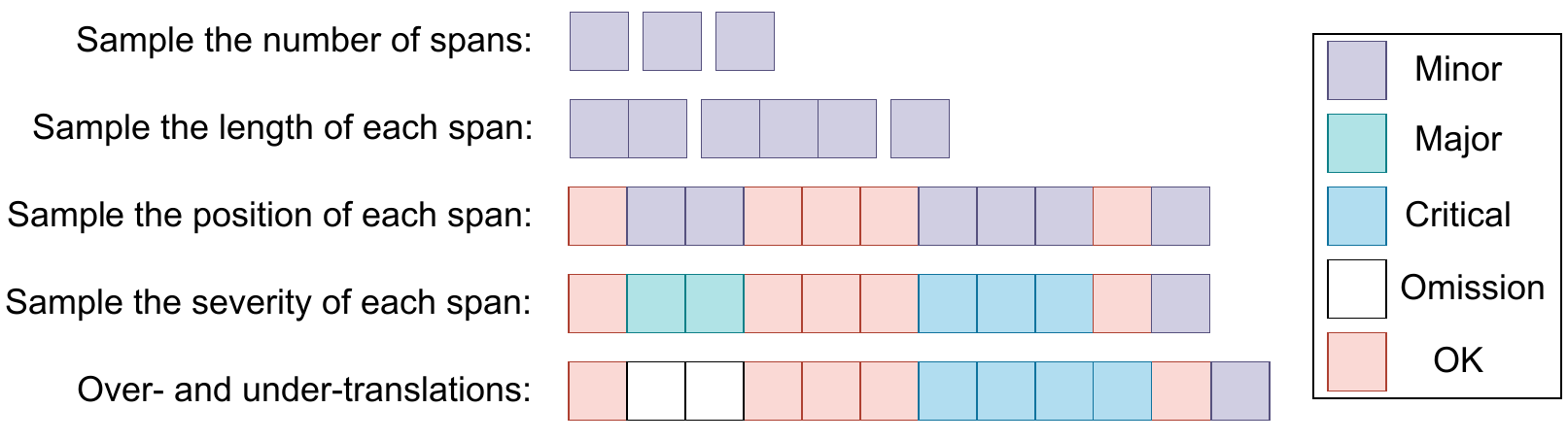}
\caption{
Illustration of the pseudo MQM data method \cite{geng2022njunlp}.  
The word-level tags of this pseudo translation are annotated as ``OK BAD OK OK BAD BAD BAD BAD OK BAD'' and the MQM score is -0.6.
}
\label{fig:mqm}
\end{figure*}

Given a parallel pair $(X, Y)$, we corrupt the reference $Y$ as shown in figure \ref{fig:mqm}:
\begin{itemize}
    \item We sample the number of spans $t$ according to the distribution of WMT2022 QE EN-DE valid set \cite{zerva-etal-2022-findings}.
    \item According to the distribution of WMT2022 QE EN-DE valid set, we sample the length of each span $n_i$ one by one to ensure that the total length is less than reference length $n$. 
    \item We randomly sample the start indices for $i$-th span in $[\text{EOL}_i,n-\sum_{j=i}^{t}n_j]$ to ensure each span lie in the sentence, where $\text{EOL}_i$ is the end indices of last span ($\text{EOL}_0=0$).
    \item We sample the severity of each span according to the distribution of a WMT2022 QE EN-DE valid set. 
    \item We randomly insert or remove some tokens in each span to simulate over- and under-translations.
    \item We tag tokens on the right of the omission errors and tokens that are not aligned with reference tokens as ``BAD''. 
    The rest tokens are tagged as ``OK''. We calculate the MQM score using Eq. \ref{eq:mqm} based on the sampled severities.
\end{itemize}
\subsubsection{Fixing}
To generate pseudo translations, we replaced these error tokens with the ``mask'' symbol and sampled these tokens with neural machine translation (NMT) model~\cite{transformer} or translation language model (TLM)~\cite{tlm}. 
For the NMT model, we generate these error tokens from left to right with teacher forcing, while the TLM model generates these tokens parallel.
To simulate errors of different severities, we sample tokens with lower generation probabilities for graver pseudo errors.
To generate diverse pseudo translations, we random sample one of the tokens with the top $k$ generation probability as the error token.
In practical, we use $k=2, 10, 100$ for minor, major, and critical errors, respectively.

\subsection{Pre-training and Fine-tuning}
\subsubsection{QE Model}
Since the pre-train models significantly improve MT evaluation performance~\cite{rei2022cometkiwi,zerva-etal-2022-disentangling}, we use the XLMR large model ($f$) as the model backbone. To obtain the features conditioned on source sentences, we input the concatenation of source sentences and translations:
\begin{align}
    H_X, H_{\hat{Y}} = f(X, \hat{Y}).
\end{align}
Then, we average the representations $H_{\hat{Y}}$ of all target tokens as the sentence score representation $H_{\text{sent}}$. 
\begin{align}
    H_{\text{sent}}=\text{Average}(H_{\hat{Y}})
\end{align}
The sentence score representation passes through one linear layer and an optional activation function $\sigma$ to output the score prediction $\hat{m}$.
\begin{align}
    \hat{m} = \sigma (\text{FFN}(H_{\text{sent}})),
\end{align}
where we set $\sigma$ as the Sigmoid function or null.
We average sub-tokens’ representations as the representation of the whole word.
We input the word representations $H_{\text{word}}$ to one linear layer and softmax function to predict binary labels:
\begin{align}
    \hat{G} = \text{softmax}(\text{FFN}(H_{\text{word}})).
\end{align}

\subsubsection{QE Loss}
Following the multi-task learning framework for QE \cite{zerva-etal-2021-ist}, we joint learn the sentence- and word-level tasks. 
We use two loss functions for the sentence-level task: the margin ranking loss and the mean square error (MSE) loss.
The margin ranking loss is defined as follows:
\begin{align}
    L_{\text{Rank}} = \max (0, -r(\hat{m}^i-\hat{m}^j)+\epsilon),
\end{align}
where $\hat{m}^i$ and $\hat{m}^j$ denote the output scores of $i$-th and $j$-th translations from current batch; $r$ denotes the rank label, $r=1$ if $m^i>m^j$, $r=-1$ if $m^i<m^j$; $\epsilon$ denotes the margin, we set $\epsilon=0.03$ for all experiments.
As shown in \cite{geng2022njunlp}, the ranking loss is critical to achieving good performance.
And the MSE loss is defined as:
\begin{align}
    L_{\text{MSE}} = \text{MSE}(m, \hat{m}).
\end{align}
We use cross-entropy (CE) loss for the word-level task:
\begin{align}
    L_{\text{CE}} = \sum_{i=1}^n\text{CE}(g_i, \hat{g}_i),
\end{align}
where $\hat{g}_i$ denotes the tag predicted for $i$-th word.
The final QE loss function is the weighted sum of previous loss functions:
\begin{align}\label{j_qe}
    L_{\text{QE}} = L_{\text{CE}}+{\alpha}L_{\text{MSE}}+{\beta}L_{\text{Rank}},
\end{align} 
where ${\alpha}$ and ${\beta}$ denote the weights for different loss functions.
We use the Eq. \ref{j_qe} for both pre-training and fine-tuning.

\subsection{Ensemble}
We generate one pseudo MQM data for each parallel pair. 
We train different QE models with different pseudo MQM data and ensemble their results as the final submission.
For the sentence-level task, we calculate the z-scores of each output and the average of these z-scores as the predictions.
For the word-level task, we use QE models to output ``OK'' probabilities $P=\{p_1,p_2,\dots,p_n\}$, where $p_i$ denotes the ``OK'' probability for $i$-the word in the translation. Then, we average ``OK'' probabilities and set a threshold $\epsilon_{\text{BAD}}$ to decide whether the word is ``BAD'':
\begin{align}
    \hat{g_i} = \begin{cases}
  \text{OK} & \text{ if } p_i > \epsilon_{\text{BAD}} \\
  \text{BAD} & \text{ if } p_i \leq \epsilon_{\text{BAD}}
\end{cases}
\end{align}

\subsection{Sub-task (ii)}
To unite the word-level sub-task and fine-grained
error span detection sub-task, we propose a simple
method that covert the word-level outputs to fine-grained error span results.
Based on the ensemble ``OK'' probabilities, we set two thresholds $\epsilon_{\text{major}}$ and $\epsilon_{\text{minor}}$. 
Then, we can output the fine-grained error tags $S=\{s_1,s_2,\dots,s_n\}$, where $p_i$ as follows:
\begin{align}
    \hat{s_i} = \begin{cases}
  \text{OK} & \text{ if } p_i > \epsilon_{\text{minor}} \\
  \text{Minor} & \text{ if } \epsilon_{\text{Major}} < p_i \leq \epsilon_{\text{Minor}} \\
  \text{Major} & \text{ if } p_i \leq \epsilon_{\text{Major}}
\end{cases}
\end{align}
Finally, we regard consecutive error tokens as a whole span and take the worst severity of error tokens as the span severity. 
As recommended by the reviewer, we also try to take the majority category as the span severity. However, we found that only one prediction changed from``major'' to ``minor". 
That may be because the task is imbalanced and there are more ``major'' errors.
As a result, this strategy achieves the same F1-score as the previous one.

\section{Experiments}
\subsection{Implementation Details}
We use parallel data from the WMT translation task to generate the pseudo MQM data.
We use the WMT2022 QE EN-DE dataset and the WMT2022 Metric EN-DE dataset for fine-tuning.
We also incorporate the post-editing annotation EN-DE datasets (WMT17, 19, and 20) to warm up the QE model.

We implement our system based on the NJUQE framework, which is built on the Fairseq(-py)~\cite{fairseq} toolkit.
We use NVIDIA V100 GPUs to conduct our experiments.
To search the hyper-parameters, we utilize the grid search method.
All experiments set the random seed as 1. We set $\alpha=1$ and $\beta=1000$ for both pre-training and fine-tuning. When pre-training, we use four GPUs. We set the learning rate to 1e-5, the maximum number of tokens in a batch to 1400 and update
the parameters every four batches. We evaluate the model every 600 updates and perform early stopping if the validation performance does not improve for the last ten runs. When fine-tuning, we use one GPU. we set the learning rate to 1e-6, the maximum number of sentences in a batch to 20. We evaluate the model every 300 updates and perform early stopping if the validation performance does not improve for the last ten runs.

\subsection{Results}
We achieve the best results on EN-DE for both word-level and fine-grained error span detection sub-tasks with an MCC of 29.7 (+4.1 than the second best system) and F1 score of 28.4 (+1.1) respectively. 
We rank 2nd place on sentence-level sub-tasks with a Spearman score of 47.9 (-0.4 than the best system).
\section{Analysis}
In this section, we show some key hyper-parameters that improve the performance.
\subsection{The normalize function $\sigma$}
Although the MSE loss improves sentence-level performance, we need to avoid the over-fitting of score predictions.
We set the normalize function $\sigma$ as the sigmoid function to provide smooth gradients.
As shown in table \ref{tab:normalize function}, we achieve better sentence-level performance by using the sigmoid function.
\begin{table}[t]
    \centering
    \begin{tabular}{c|c}
    \hline
       $\sigma$  & Spearman \\
       \hline
       w/o $\sigma$  & 50.02 \\
        sigmoid & 52.41\\
        \hline
    \end{tabular}
    \caption{Results on the validation set of WMT2022 QE EN-DE task with different normalize function $\sigma$.}
    \label{tab:normalize function}
\end{table}

\subsection{Dropout Rate of the Output Layers}
We also use the dropout method \cite{dropout} on the output layers to avoid over-fitting. 
Table \ref{tab:dropout} shows that the QE model obtains better performance when we set the dropout rate as 0.2.
\begin{table}[t]
    \centering
    \begin{tabular}{c|c}
    \hline
       Dropout Rate  & Spearman \\
       \hline
       0  & 52.41 \\
        0.1 & 52.93\\
        0.2 & 53.11\\
        0.3 & 52.15\\
        \hline
    \end{tabular}
    \caption{Results on the validation set of WMT2022 QE EN-DE task with different dropout rate.}
    \label{tab:dropout}
\end{table}

\section{Conclusion}

We present NJUNLP’s work to the WMT 2023
Shared Task on Quality Estimation. 
In this work, we generate pseudo MQM data using parallel data.
We pre-train the XLMR large model on pseudo MQM data, then fine-tune it on real QE data.
At both stages, we jointly learn sentence-level scores and word-level tags. 
Empirically, we conduct experiments to find the key hyper-parameters that improve the performance.
Technically, we propose a simple method that covert the word-level outputs to fine-grained error span results.
Overall, our models achieved the best results in English-German for both word-level and fine-grained error span detection sub-tasks by a considerable margin.
\section*{Acknowledgements}
We would like to thank the anonymous reviewers for their insightful comments. Shujian Huang is the corresponding author. This work is supported by National Science Foundation of China (No. 62376116, 62176120), the Liaoning Provincial Research Foundation for Basic Research (No. 2022-KF-26-02).
\bibliography{custom}
\bibliographystyle{acl_natbib}
\end{document}